\documentclass{article}

\usepackage[utf8]{inputenc}
\usepackage{amsmath, amsthm, amssymb, amsfonts, dsfont, mathrsfs, dirtytalk, tikz-cd, adjustbox, url, nicefrac, booktabs,array, graphicx,todonotes,longtable, tabu,bm, todonotes,comment,multirow,tabularx,epsfig,parskip,cite}
\usepackage[font=footnotesize]{caption}
\usepackage[T1]{fontenc}    
\usepackage[english]{babel}
\usepackage{hyperref}
\usepackage[capitalise]{cleveref}

\usepackage[preprint, nonatbib]{neurips_2024}

\usepackage[utf8]{inputenc} 
\usepackage[T1]{fontenc}    
\usepackage{hyperref}       
\usepackage{url}            
\usepackage{booktabs}       
\usepackage{amsfonts}       
\usepackage{nicefrac}       
\usepackage{microtype}      
\usepackage{xcolor}         

\title{Toward Universal and Interpretable World Models for Open-ended Learning Agents}

\author{%
  Lancelot Da Costa \\
  VERSES AI Research Lab\\
  \texttt{lance.dacosta@verses.ai} \\
}

\begin{document}

\maketitle


\begin{abstract}
We introduce a generic, compositional and interpretable class of generative world models that supports open-ended learning agents. This is a sparse class of Bayesian networks capable of approximating a broad range of stochastic processes, which provide agents with the ability to learn world models in a manner that may be both interpretable and computationally scalable. This approach integrating Bayesian structure learning and intrinsically motivated (model-based) planning enables agents to actively develop and refine their world models, which may lead to developmental learning and more robust, adaptive behavior.
\end{abstract}

\section{Introduction}

A feature of open-ended learning agents is the ability to gradually understand wide-ranging, complex worlds. Computational accounts of cognitive development in humans and animals frame this process as approximate Bayesian inference over spaces of Bayesian networks, where inferences are continuously refined thanks to sensory data arriving from active (intrinsically motivated) interactions with the environment \cite{ullman_bayesian_2020}.
Current modelling approaches to development, while impressive, face scalability challenges, due to the explosive nature of inferring over \emph{all} possible Bayesnets 
\cite{tsividis_human-level_2021,pouncy_inductive_2022}. See Appendix \ref{app: b} for more context. Thus, a crucial challenge to enable developmental agents is finding Bayesnet spaces that can express a wide range of worlds, while being coarse enough to efficiently be searched.

\section{A generic, interpretable and agentic class of generative models}
\label{sec: introducing gen mod class}

Desiderata for this Bayesnet class include being: \emph{(i)} \textbf{Sufficiently expressive} to be able to approximately express relevant naturalistic, dynamic interactions between agent and environment. \emph{(ii)} \textbf{Sufficiently coarse} so that inference on this space can be made computationally tractable. Each Bayesnet in this class should be: \emph{(iii)} \textbf{Interpretable} so that the agents' understanding and ensuing behaviour can be easily understood from the Bayesnets it entertains. \emph{(iv)} \textbf{Support fast action, perception and learning}. We summarise a candidate class of Bayesnets satisfying requirements \emph{(i)-(iv)} in Appendix \ref{app: c}, extending earlier proposals \cite{friston_supervised_2023}.

\section{Discussion}

The expressivity of this class lies in its ability to represent complex stochastic processes, including those with both discrete and continuous states. By utilizing hierarchical structures, these models can capture both high-level abstractions and fine-grained details of the environment, allowing agents to form robust and generalizable representations. The interpretability of these models stems from their sparsity and encoding of causal relationships---this provides insights into the agent's decision-making, facilitating more transparent and trustworthy AI systems. This class of Bayesnets has been used to model video from raw pixels and sound files, and agents that plan from pixels \cite{friston_pixels_2024}. Future work should seek the limits of this approach, including its scalability and ability to express relevant agent-environment interactions.

\bibliographystyle{unsrt}
\bibliography{bib}

\appendix

\section{Current challenges in developmental agents}
\label{app: b}

Engineering agents that scalably learn models of the world remains a relatively open challenge \cite{scholkopf_toward_2021}, despite the fact that cognitive development of humans and animals is well-studied within computational cognitive science \cite{ullman_bayesian_2020}.
Several strands of work in computational cognitive science have converged to the idea that the developmental process is a process of approximate Bayesian inference about explanations for the world (i.e. Bayesian networks) and how the agent interacts with it, where inferences are gradually refined by actively sampling new, informative data (e.g. through intrinsic motivation) \cite{ullman_bayesian_2020,gopnik_theory_2004,goodman_learning_2011,de_tinguy_exploring_2024,friston_pixels_2024}. Engineering this simple picture is difficult as the search space of possible explanations for the world explodes combinatorially in the number of variables being modelled \cite{weisstein_acyclic_nodate}. To illustrate the problem further, we summarise the current theory-based reinforcement learning work \cite{tsividis_human-level_2021,pouncy_inductive_2022}, which simulates developmental agents by combining Bayesian structure learning and intrinsically motivated (model-based) planning.

In foundational papers \cite{tsividis_human-level_2021,pouncy_inductive_2022}, an agent maintains Bayesian beliefs about probabilistic programs (which encode Bayesnets) that might explain the world. The agent gradually refines refines its inferences by actively seeking new data through a mixture of exploration and exploitation (i.e. intrinsic motivation \cite{schmidhuber_formal_2010,deci_intrinsic_1985}). The authors deployed this agent in a suite of simplified Atari games, and found not only that their agent averaged human learning efficiency across the games (after comparing with data from human participants), but the agents' learning trajectories were relatively similar to that of humans. This work serves as a proof of concept that combining inference about the structure of the world with intrinsically motivated model-based planning can achieve relatively human-like learning and behaviour.

The fundamental limitation of existing theory-based RL work is that their agents \cite{tsividis_human-level_2021,pouncy_inductive_2022} consider a search space of explanatory hypotheses about the world that is the whole set of programs (up to a certain length) that can be generated from the code grammar generating the data generating process. This is an extremely large search space even for the simplified Atari environments their agents face and highlights the current limitations of this work: 1) In complex environments, the space of programs that can be generated from the code grammar generating the data-generating process may be far too large to be searchable, 2) in real environments the modeler does not know the generative process and cannot easily form a space of candidate explanations that contains the data generating process.

Thus a fundamental question is what might be a `universal' set of primitives and compositional rules to produce a space of candidate explanations for the world (i.e. Bayesnets) satisfying requirements \emph{(i)-(iv)} in Section \ref{sec: introducing gen mod class}.
Indeed, there is already a tension between requirements \emph{(i)-(2)} and a significant difficulty lies in striking the right balance. Asking what a universal space of Bayesnets might look like, we first consider the existing literature: Spaces of \textbf{probabilistic programs }are easily made extremely expressive, but it is not clear how to do so while keeping them coarse enough for inference to remain tractable. Probabilistic programs are not always easily interpretable, and, barring specific assumptions, do not support efficient perception and learning, as Bayesian inference over states and parameters may require sampling. One example of probabilistic programs that might satisfy these requirements---to a first approximation---are \textbf{hierarchical discrete and continuous state partially observed Markov decision processes} (POMDPs) \cite{astrom_optimal_1965,barto_reinforcement_1992}. 

This space of Bayesnets is very expressive in being able to reproduce a wide variety of behavioural simulations and empirical data. For example, nearly all modeling work in the active inference literature alone, which spans nearly two decades, employed Bayesnets that are built by hierarchically stacking these two types of layers \cite{da_costa_active_2020,parr_active_2022,smith_step-by-step_2022,lanillos_active_2021}. The resulting Bayesnets support fast action, perception and learning where inference about states and parameters is done through fast variational inference procedures \cite{da_costa_active_2020,parr_neuronal_2019,heins_collective_2024,parr_active_2022,friston_action_2010}, which have a degree of biological plausibility in being able to reproduce a wide range of features from real neural dynamics, e.g. \cite{friston_hierarchical_2008,friston_active_2017-1,isomura_canonical_2022,isomura_experimental_2023}. Barring the use of neural networks for expressing non-linearities in these layers \cite{mazzaglia_free_2022}, each of them furnishes an interpretable model of dynamics.


\section{Details on generic, interpretable and agentic class of Bayesnets}
\label{app: c}

The Bayesnet class we consider here as potential explanations for the world is built from a simple set of primitives and compositional rules. We claim that it satisfies the requirements \emph{(i)-(iv)} from Section \ref{sec: introducing gen mod class}. \textit{(i) }Its expressivity lies in its ability to represent complex stochastic processes, including those with both discrete and continuous states. \textit{(ii) }Its sparsity stems from significant inductive biases on the structrures of the Bayesnets being considered. \textit{(iii) }The interpretability of each Bayesnet in this class stems from
their sparsity as well as encoding of causal relationships. Finally, and as we will see, these Bayesnets support fast action, perception and decision-making \textit{(iv)}. 

The proposed Bayesnet class is built by hierarchically assembling a set of basic structural modules satisfying requirements \textit{(iii)-(iv)}, which together, express a wide range of dynamic agent-environment interactions. In what follows, we detail each of these building blocks developed specifically to express a large class of stochastic processes on either discrete or continuous states.

\subsection{Discrete dynamics}

\begin{figure}
    \centering
    \includegraphics[width=\linewidth]{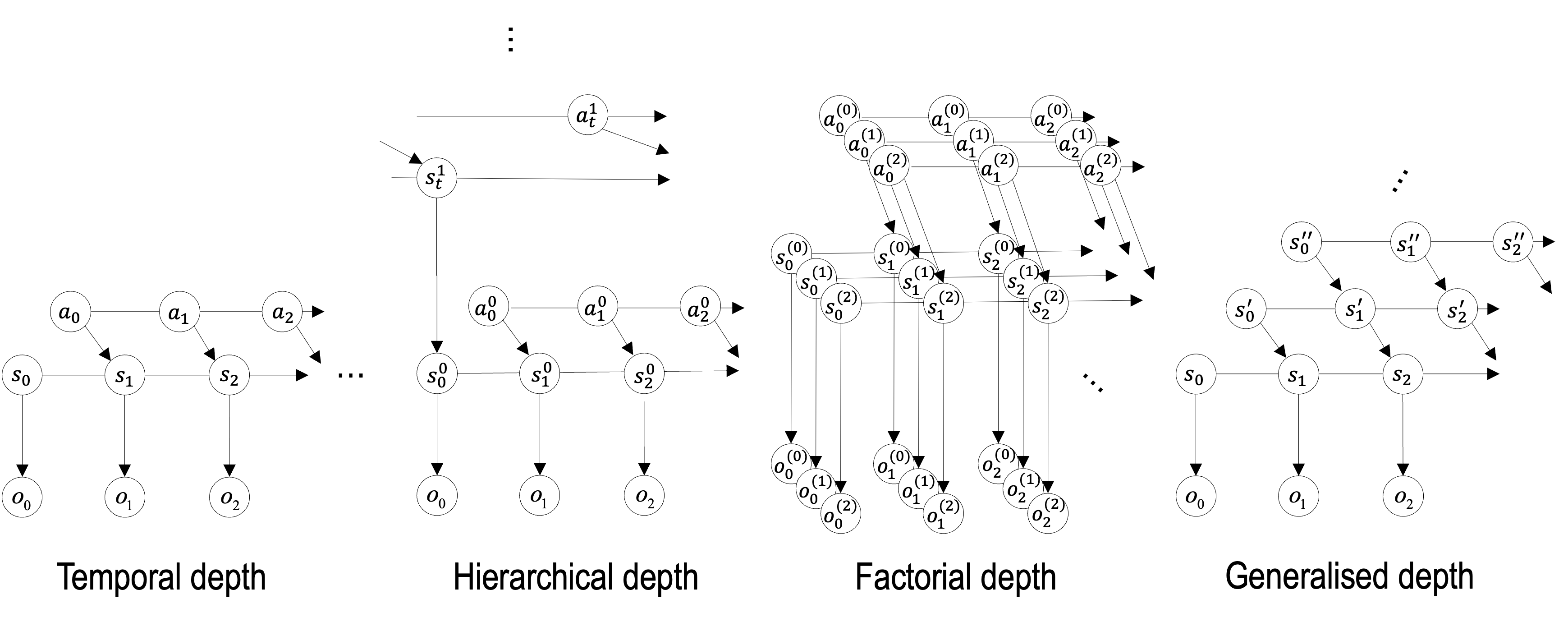}
    \caption{\textbf{Space of discrete-state Bayesian networks.} The basic module for expressing discrete-state agent-environment interactions is a partially observed Markov decision process (POMDP). This module can have an arbitrary finite temporary horizon (i.e. temporal depth). Several such modules can be stacked atop each other finitely many times (i.e. hierarchical depth), thereby expressing multi-scale semi-Markovian latent dynamics. We can specify multiple co-evolving factors (i.e. factorial depth; e.g. position and colour of an object). In any given layer, a number of auxiliary (i.e. generalised) states can be added, accounting for velocity, acceleration, and higher orders of motion in the latent states (i.e. generalised depth) to express semi-Markovian processes \cite{friston_supervised_2023}. In each of these layers, the highest generalised states may or may not be actions denoted by $a$ (cf. controllable states \cite{friston_supervised_2023}) while all other states are uncontrollable (i.e. part of the environment) denoted by $s$.
    The controllable and uncontrollable states cause observations denoted by $o$. The parameters of these Bayesian networks (not shown) correspond to the parameters of the causal maps as well as the initial distributions over states \cite{da_costa_active_2020}. These parameters as well as the graphical structure of these Bayesnets to be inferred from data (e.g. past actions and observations). The resulting approximate posterior belief then informs action, which follows a mixture of goal-seeking and information gathering (i.e. intrinsic motivation) imperatives. Please see \cite{friston_supervised_2023,friston_active_2023} for more details on temporal, hierarchical and factorial depth---and examples of controllable versus uncontrollable states.}
    \label{fig: discrete}
\end{figure}

Markov processes are a fairly ubiquitous class of stochastic processes \cite{hairer_markov_2020}. All Markov processes on discrete states have simple transition dynamics that are given by linear algebra. 
When these transitions also depend on actions, we obtain Markov decision processes \cite{puterman_markov_2014}. When states are partially observed and observations depend only on the current latent state, we obtain POMDPs. We can add auxiliary latent states to those POMDPs \cite{friston_supervised_2023} (i.e. the equivalent of momentum, acceleration etc) to account for the effect of memory in the system, producing semi-Markovian POMDPs. Lastly, we can stack these layers hierarchically to express multi-scale semi-Markovian processes in latent space. In summary, hierarchical discrete POMDPs extended in this way provide a very generic class of models for agent-environment interactions on discrete states. See Figure \ref{fig: discrete} for a graphical representation of discrete extended POMDPs and their various degrees of freedom.

\subsection{Continuous dynamics}

For expressing continuous dynamics, the situation is somewhat more difficult. Repeating the construction of the discrete state-space model seems hardly possible because continuous-space Markov processes are given by linear operators in infinite (as opposed to finite) dimensional spaces \cite{bakry_analysis_2014}. A working alternative is to restrict ourselves to a more manageable but still very expressive class of processes. We can consider continuous POMDPs with latent dynamics given by stochastic differential equations (SDEs), which is another very expressive class of stochastic processes. However, for expressing a wide range of agent environment interactions, these SDEs must break detailed balance and possibly be driven by non-Markovian noise (fluctuations in the wind or in the ocean)\cite{parr_computational_2021}. There is a remarkably expressive class of SDEs supporting non-linearities, non-Markovian noise and broken detailed balance---that is, many times differentiable stochastic differential equations---for which POMDPs with these latent dynamics support fast and biologically plausible update rules for action, perception and learning \cite{heins_collective_2024,parr_active_2022,friston_hierarchical_2008}.
These continuous POMDPs form a very expressive space of continuous-state Bayesian networks by varying the temporal, hierarchical, factorial and generalised depth as in Figure \ref{fig: discrete}.

\begin{figure}
    \centering
    \includegraphics[width=0.6\linewidth]{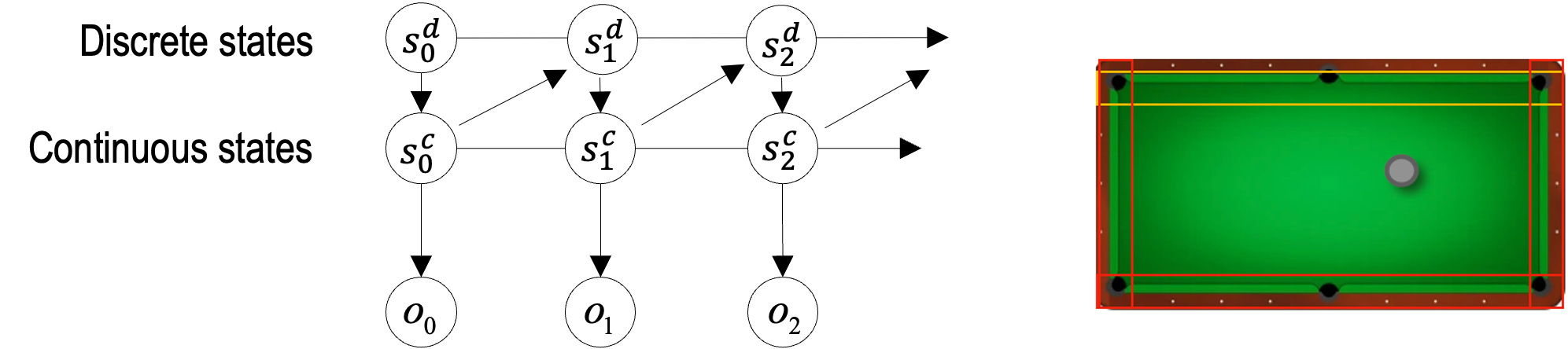}
    \caption{\textbf{Recurrent switching linear dynamical systems (rsLDS).} 
    This figure illustrates rsLDS as a way of interpretably parameterising fairly arbitrary stochastic differential equations (SDEs) as a switching linear SDE. Consider for example a generative model used to play pool. This must be able to express the non-linear trajectories in a game of pool. The behaviour of a ball on a pool table depends on whether the ball is by one of the four walls (i.e. bouncing), or not. In each of these sectors the dynamic of the ball is captured by a simple dynamical equation. The rsLDS \cite{linderman_bayesian_2017} is a simple generative model that is able to express these kinds of trajectories, where a (switching) discrete state expresses in which of these sectors the ball is. This discrete state furnishes a linear drift and volatility to the linear SDE describing the continuous motion of the ball (cf. continuous states), which yields continuous observations. Note the so-called recurrent connection (i.e. causal map) from continuous to discrete states; this connection enables the continuous dynamic to influence the discrete switching: if the continuous trajectory of the ball collisions with a wall the discrete state switches so that the continuous dynamic of the ball changes course. The rsLDS layer may be supplemented with active states, acting on the discrete latent states, thereby emulating a continuous partially observable Markov decision process. This continuous POMDP can easily be extended by varying the temporal, hierarchical, factorial and generalised depth as in Figure \ref{fig: discrete}, furnishing a generic model of continuous dynamics. Please see \cite{linderman_bayesian_2017} for more details on the current rsLDS architecture.}
    \label{fig: rslds}
\end{figure}

One important challenge remains: to parameterise the non-linearities in these POMDPs (e.g. flows of SDEs) without sacrificing interpretability, and learn these parameterisations from data. A promising approach is to express non-linear SDEs with recurrent switching linear dynamical systems (rsLDS; see Figure \ref{fig: rslds}) \cite{linderman_bayesian_2017}; that is, switching mixtures of linear SDEs, because one could use a very fine grained piece-wise linear approximation to recover arbitrary non-linearities, as necessary.
The advantage of using switching linear SDEs is that they are interpretable and afford relatively scalable exact Bayesian inference \cite{linderman_bayesian_2017}.\footnote{An alternative way of inferring non-linearities would be by parameterising them with Bayesian neural networks (i.e. neural SDEs \cite{kidger_neural_2022}), but this would likely hinder interpretability.}
However, current rsLDS architectures are restricted to approximating the dynamics of non-linear so-called 'diffusion' SDEs discretised with an Euler scheme \cite{linderman_bayesian_2017}, which do not feature non-Markovian noise by definition. Looking forward, it seems desirable to extend the rsLDS architecture to express SDEs with more arbitray noise signals, perhaps by adding generalised coordinates (velocity, acceleration, and higher orders of motion etc) \cite{heins_collective_2024,friston_hierarchical_2008}. This would entail introducing generalised depth into the current rsLDS layer. Doing this should furnish a very expressive yet sparse class of Bayesnets for continuous-state dynamics that satisfies the basic requirements \textit{(i)-(iv)}.

\subsection{Hierarchical mixed dynamics}

Stacking hierarchies of discrete layers atop hierarchies of continuous layers yields mixed generative models that can express rich non-linearities and dynamics at several levels of abstraction. Despite the absence of traditional neural networks in these hierarchies, one can think of them as a neural network where the layers are discrete and continuous POMDPs and the computations are of efficient approximate Bayesian inference. Hierarchies of these layers may be interpretable as they represent nested processes operating at different timescales. These hierarchical structures are compatible with views of the brain as entertaining discrete-state low-dimensional abstract dynamics that condition the high-dimensional continuous representations closer to sensory input  \cite{parr_discrete_2018,friston_graphical_2017}.



\end{document}